\documentclass[letterpaper, 10 pt, conference]{ieeeconf}  

\pdfoutput=1
\IEEEoverridecommandlockouts                              

\usepackage{graphics} 

\usepackage{times} 
\usepackage{amsmath} 
\usepackage{amssymb}  
\usepackage{algpseudocode, algorithm}
\usepackage[keeplastbox]{flushend} 
\usepackage{caption}
\usepackage{subfig}
\usepackage{graphicx}
\usepackage{multirow}
\usepackage{bm}
\usepackage{booktabs}
\usepackage{array}
\usepackage{gensymb} 
\usepackage{threeparttable}
\usepackage[dvipsnames]{xcolor}
\definecolor{red}{rgb}{1, 0, 0}
\definecolor{green}{rgb}{0, 1, 0}
\definecolor{grey}{rgb}{0.8,0.8,0.8}
\definecolor{yellow}{rgb}{1,1,0}
\newcolumntype{C}[1]{>{\centering}m{#1}}

\title{\LARGE \bf
Constrained Heterogeneous Vehicle Path Planning \\
for Large-area Coverage
}

\author{Di Deng$^{1}$, Wei Jing$^{2}$, Yuhe Fu$^{1}$, Ziyin Huang$^{1}$, Jiahong Liu$^{1}$ and Kenji Shimada$^{1}$
\thanks{$^{1}$Faculty of Mechanical Engineering, Carnegie Mellon University, 5000 Forbes Ave, Pittsburgh, PA 15213, USA. {\tt\small dengd@andrew.cmu.edu}}%
\thanks {$^{2}$A*STAR Artificial Intelligence Initiative (A*AI); and Dept. of CS, IHPC, A*STAR; 1 Fusionopolis Way, Singapore 138632 {\tt\small jing\_wei@ihpc.a-star.edu.sg}}%
}

\begin{document}
\maketitle

\noindent\begin{abstract}
There is a strong demand for covering a large area autonomously by multiple UAVs (Unmanned Aerial Vehicles) supported by a ground vehicle. Limited by UAVs' battery life and communication distance, complete coverage of large areas typically involves multiple take-offs and landings to recharge batteries, and the transportation of UAVs between operation areas by a ground vehicle. In this paper, we introduce a novel large-area-coverage planning framework which collectively optimizes the paths for aerial and ground vehicles. Our method first partitions a large area into sub-areas, each of which a given fleet of UAVs can cover without recharging batteries. UAV operation routes, or trails, are then generated for each sub-area. Next, the assignment of trials to different UAVs and the order in which UAVs visit their assigned trails are simultaneously optimized to minimize the total UAV flight distance. Finally, a ground vehicle transportation path which visits all sub-areas is found by solving an asymmetric traveling salesman problem (ATSP). Although finding the globally optimal trail assignment and transition paths can be formulated as a Mixed Integer Quadratic Program (MIQP), the MIQP is intractable even for small problems. We show that the solution time can be reduced to close-to-real-time levels by first finding a feasible solution using a Random Key Genetic Algorithm (RKGA), which is then locally optimized by solving a much smaller MIQP. 

\end{abstract}

\section{Introduction}
\begin{figure}[t!]
     \centering
     \subfloat[Farmlands in Niigata, Japan]{\includegraphics[width=0.43\linewidth]{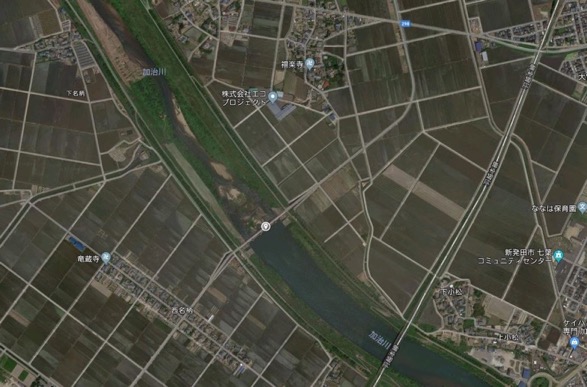}\label{fig:farm_land}}
     \hspace{5pt}
     \subfloat[A UAV spraying pesticide \cite{dji}]{\includegraphics[width=0.52\linewidth]{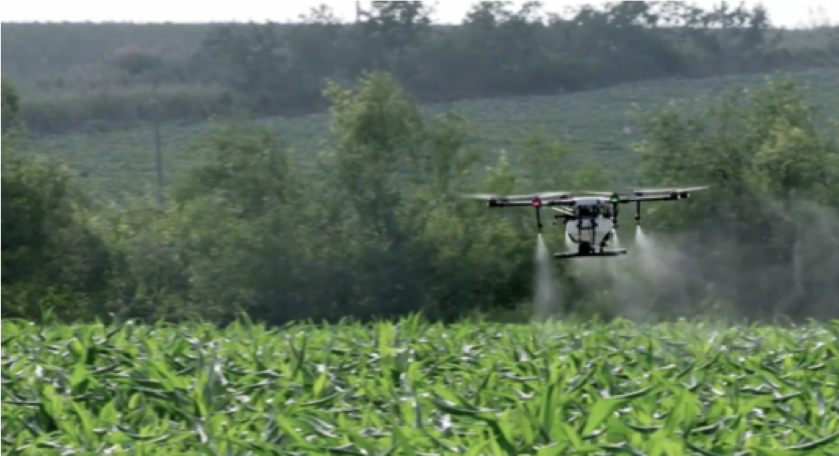}\label{fig:uav_coverage}}
     \caption{UAV for agricultural application}
     \label{fig:figure_1}
\end{figure}

\noindent 
As the agricultural industry in east Asia suffers from increasingly severe labor shortage, the need to automate away as much work as possible is pressing. As a result of this automation trend, the market for autonomously spraying pesticides and fertilizer with Unmanned Aerial Vehicles (UAVs) have been growing quickly (Fig. \ref{fig:figure_1}). In fact, the market for pesticide-spraying UAVs will expand fifteen times from 2016 to 2022, according to a survey conducted by Seed Planning, Inc. \cite{xiongkui2017recent}.

To cover large farmland, it is necessary to deploy a team of multiple ground and aerial vehicles because the area which can be covered with a single take-off and landing by one UAV is limited by its battery life and maximum communication distance. In such deployments, a ground vehicle is used to recharge UAV batteries and monitor the UAV fleet. 

Coverage path planning (CPP) is a class of algorithms that find paths for one or multiple robotic agents that completely cover/sweep a given task area. It is an essential component for applications such as room floor sweeping, coastal area inspections \cite{karapetyan2018multi}, 3D reconstruction of buildings \cite{jing2016view}, and, of course, autonomous pesticide and fertilizer spraying (also known as crop dusting) \cite{dji, faiccal2017adaptive}.

As we will detail in the rest of this paper, autonomous crop dusting by a fleet of UAVs and a ground vehicle poses interesting and unique constraints that existing CPP planners cannot handle effectively. Our contribution is a novel, fast planner for crop dusting which provides locally optimal paths that satisfy the aforementioned application-specific constraints. 


The rest of the paper is structured as follows: after related work is reviewed in Sec. \ref{sec:work}, Sec. \ref{sec::prop_method} gives a mathematical description of the heterogeneous vehicle coverage problem and our proposed planning framework; Sec. \ref{sec:result} takes Niigata's (a prefecture in Japan) farmland (Fig. \ref{fig:farm_land}) as an example to demonstrate the effectiveness of our algorithm. 

\section{Related Work\label{sec:work}}
\begin{figure*}[t!]
    \vspace{5pt}
    \centering
    \includegraphics[width=\linewidth]{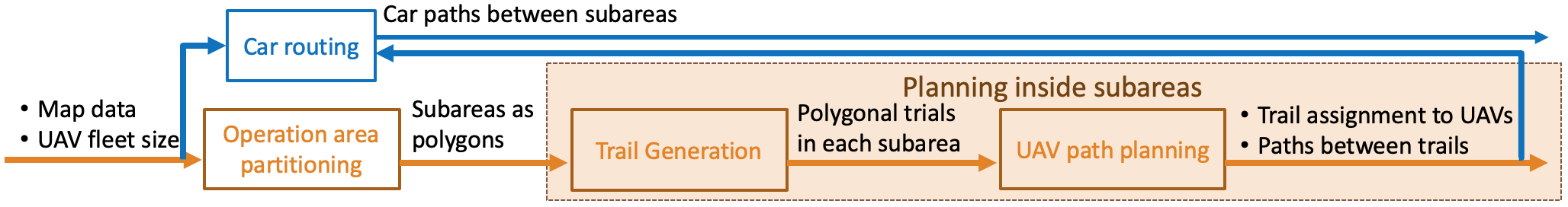}
    \caption{Overview of the proposed coverage planning framework}
    \label{fig:big_picture}
\end{figure*}
\noindent 
According to surveys conducted by Galceran and Carreras \cite{galceran2013survey} and Choset \cite{choset2001coverage}, CPP algorithms can be classified into two broad categories: cellular decomposition and grid-based methods. Cellular decomposition partitions general non-convex task areas, typically in the form of 2D polygons, into smaller sub-regions with nice properties such as convexity \cite{choset2000coverage, maza2007multiple, balampanis2017area, bast2000area}. In the partitioned sub-regions, coverage path patterns can then be easily generated using zig-zag or contour offset \cite{bormann2018indoor}. This technique can be readily extended to multiple agents by adding a planning step that assigns sub-regions to agents in the fleet. Although existing partitioning techniques have implicit notions of optimality such as path efficiency (e.g. total turning angles) \cite{choset2000coverage}, they do not jointly optimize multiple objectives, which is important for finding sub-regions in which a fleet of UAVs can efficiently operate. In addition, despite specialized CPP algorithms that can handle specific types of constraints \cite{isler2018coverage, richards2015user}, existing techniques are not good at finding optimal paths which also satisfy more general and complex constraints that are crucial for crop dusting.

On the other hand, grid-based methods, as the name suggests, discretize the task area into uniform grids. Cells in the grid can be interpreted as a nodes of a graph, so that graph search methods such as Traveling Salesman Problem (TSP) or vehicle routing problem (VRP) can be applied to find paths that visit all nodes \cite{moravec1985high, barrientos2011aerial}. However, the discretization makes it difficult to enforce the constraint that pesticides should not be sprayed in non-farmland areas, especially when the boundary of such areas lies inside cells. This can be somewhat relieved by increasing the resolution of the grids, but doing so artificially inflates the problem size and increases solving time \cite{brown2017coverage}.

Primarily due to its Turing completeness, Mixed-Integer Programming (MIP) has been applied to many flavors of planning problems \cite{jing2016sampling, maini2018cooperative, deng2018heterogeneous}. However, MIPs have exponential worst-case complexity and typically do not scale well in practice. 

In this paper, we propose a coverage planning framework that both capitalizes on the expressiveness of MIPs to satisfy constraints, and accelerates MIPs by finding good, feasible initial guesses using Genetic Algorithms (GA). We also propose a GA-based partitioning method that optimizes multiple objectives.

\section{Overview \label{sec::prop_method}}
\noindent Patches of farmland are abstracted into (possibly non-convex) polygons in $\mathbb{R}^2$. The task is to design paths for all UAVs and ground vehilces that
\begin{itemize}
    \item completely cover the given polygons,
    \item cannot fly during spraying above designated areas inside the farmland (\textit{obstacles}), such as warehouses or pump stations,
    \item minimize the UAV flying distances,
    \item respect UAV battery life, and
    \item ensure the ground vehicle stay within the communication radius of all UAVs.  
\end{itemize}

Our proposed solution as shown in Fig. \ref{fig:big_picture} is composed of the following four sequential steps.
\subsubsection{Partitioning operation area}
\label{sec:partition_requires}
As the entire target area is too large to be covered without recharging, the target area is first partitioned into smaller pieces, or \textit{sub-areas}. The size of each sub-area is limited by the number of UAVs in the fleet, the UAV's battery life, and the communication radius of UAVs. 
\subsubsection{Trail generation} In each sub-area, multiple UAV flying paths (\textit{trails}) which completely covers the sub-area are generated based on the UAV's coverage width (analogous to a camera's field of view).
\subsubsection{UAV path planning within sub-areas} Trails need to be assigned to individual UAVs in the fleet. Moreover, each UAV's assigned trails need to be connected. The assignment and the connecting paths are found by solving an optimization which minimizes the flight distance of the entire fleet.
\subsubsection{Car routing between sub-areas} After covering one sub-area, the UAV fleet returns to a ground vehicle (car) to get recharged and transported to the next sub-area. This step finds a path that visits all sub-areas and minimizes the distance traveled by the car. 

\section{Partitioning Operation Area}
\noindent First, the boundaries of farmland and the obstacles are extracted from Google Maps, as shown in Fig. \ref{fig:farm_land}. The farmland polygons, denoted by ($\mathrm{\rho}$), are then split into subareas ($\rho_1, \rho_2, \dots, \rho_n$). As stated in in Sec. \ref {sec:partition_requires}, the size of the sub-area is constrained by the number of UAVs in a fleet, $K$, their maximum pesticide spraying area, $A_{max}$, and maximum communication distance between the fleet and remote controller $L$. Another objective of partitioning is to split the original polygon into "round" rather than "skinny" sub-areas, so that it is easier for UAVs to stay close to the ground vehicle. 

Thus, the total number of sub-areas, $n$, equals to the area of $\rho$ divided by the maximum area that $K$ UAVs can cover, that is $n=\lceil\frac{A(\rho)}{KA_{max}}\rceil$. If the size of partitioned sub-regions are larger than the maximum area, $KA_{max}$, or the maximum radio transmission distance, we will increase the number of sub-areas until reading a feasible solution. 

A Genetic Algorithm (GA), similar to Algorithm \ref{algo:MVRP-RKGA} but with a different definition of chromosomes and fitness function, is used to find a relatively balanced division based on the area and dimension constraints. 

\subsubsection{Definition of chromosomes} 
\begin{figure}[htp!]
    \vspace{5pt}
    \centering
    \includegraphics[width=0.8\linewidth]{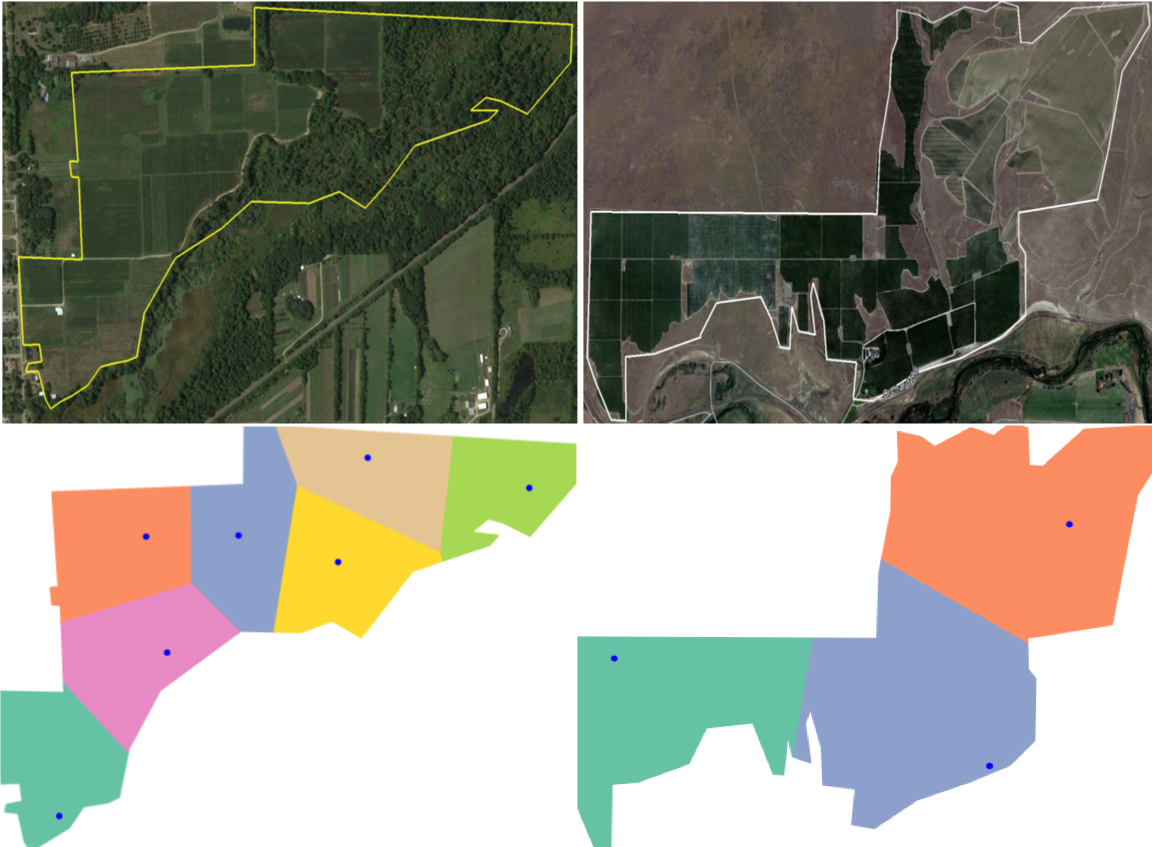}
    \caption{Field partitioning based on Voronoi diagram and GA}
    \label{fig:seed_ga}
\end{figure}

The population of GA is a 3d array $P \in \mathbb{R}^{N \times n \times 2}$, where $N$ is the number of chromosomes. $P_i[j]\in \mathbb{R}^2$ is the gene of the chromosomes. It is a 2 dimensional coordinate of a point inside farmland $\rho$. $P_i \in \mathbb{R}^{n \times 2}$ is a chromosome, representing the coordinates of a list of $n$ seed points illustrated as blue points in Fig. \ref{fig:seed_ga}.

\subsubsection{Evaluate the fitness of a chromosome} At each iteration, a partition is generated by computing the Voronoi Diagram of the seed points in a chromosome ( Fig. \ref{fig:seed_ga}). The fitness of a chromosome is defined as:
\begin{equation}
    fitness =  \frac{\omega_1}{\sigma^2(A(P_i))}+ \frac{\omega_2\mu(C(P_i))^2}{\sigma^2(C(P_i))}+\frac{\omega_3}{C(P_i)_{max}},
\end{equation}
where $\sigma^2$ is the variance, $\mu$ is the mean, $A(P_i)$ and $C(P_i)$ are the list of all areas and perimeters of sub-regions generated from Chromosome $P_i$ and $\omega_i \in \mathbb{R}$ are the weights. The first term in the fitness function minimizes the variances of the areas; the second term is heuristics for sub-areas to be more "round"; the third term minimizes the maximum perimeter among all sub-regions.

 After evaluating the fitness of all chromosomes, the ones with the highest fitness, together with some random off-springs generated by cross-over and mutation, are passed on to the next iteration until the fitness converges. The partitioning result of the proposed method is shown in Fig. \ref{fig:seed_ga}. 

\section{Trail Generation within Sub-areas \label{sec:trail_generation}}
\begin{figure}[htp]
     \centering
     \includegraphics[width=0.95\linewidth]{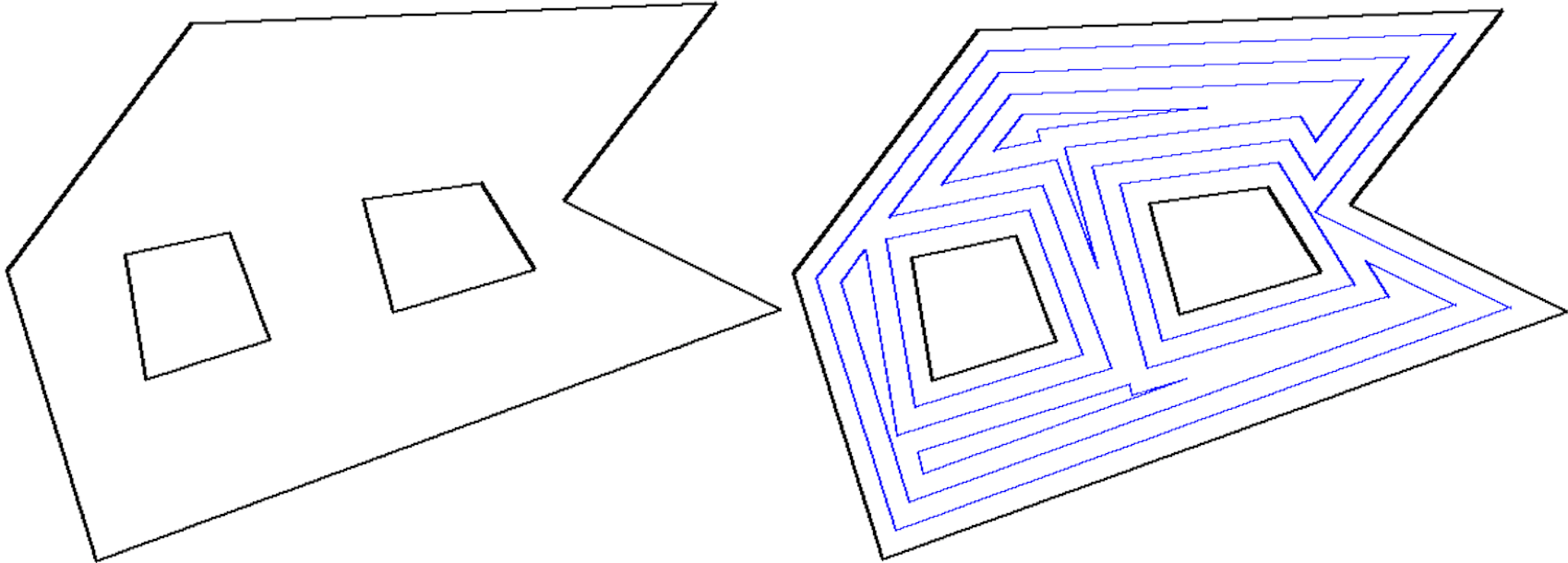}
     \caption{Mitered offset trail generation in a sub-area}
     \label{fig:loop}
\end{figure}

\noindent Mitered offset is a commonly used tool in CAM (Computer-Aided Manufacturing) software to generate tool paths \cite{huber2011computing}. As the UAVs sweep a sub-area in a similar way as a CNC (Computer Numerical Control) mill cuts profiles in a workpiece, mitered offset is employed to generate polygonal paths, or \textit{trails}, which completely cover the designated sub-area, even when the sub-area has obstacles, as shown in Fig. \ref{fig:loop}. Trails generated in this way are preferred over zig-zag paths because they contain significantly less sharp turns.

\section{Path Planning for Multiple UAVs in Sub-area \label{sec:sub_area_planning}}
\noindent After generating coverage trails for a sub-area, the next step is to assign the trails to a fleet of UAVs. An assignment, which we will call a \textit{plan}, is defined as
\begin{itemize}
    \item the sequence of trails each UAV visits, and
    \item the entry/exit points on each trail. 
\end{itemize}
Note that each trail is assigned to only one UAV. As UAVs need to fly at a fixed altitude during spraying, constraining them to fly on trails that do not cross each other significantly reduces the chances of collision.

As an example, a simple plan of a fleet of one UAV is shown in Fig. \ref{fig:trail_schematics}. In this plan, UAV 1 starts at $x_{11}$, traverses Trail 1, returns to $x_{11}$, flies to $x_{12}$ following the green dotted line, traverses trail 2, returns to $x_{12}$ and finishes the plan. We will refer to $x_{11}$ and $x_{12}$ as the \textit{access points} of Trail 1 and Trail 2, respectively. 

A plan also needs to satisfy the following requirements:
\begin{itemize}
    \item each trail is assigned to exactly one UAV, and
    \item each UAV needs to complete its assigned trails within battery constraint. 
\end{itemize}

The assignment should also minimize the total flight distances of all UAVs in the fleet. 

In this section, we show that the search for the optimal assignment can be formulated as an MIQP. However, the full MIQP has too many binary variables, thus becomes intractable for practical (moderately large) problems. To circumvent this limitation, we first search for a feasible assignment using Genetic Algorithm, which fixes most of the integer variables. A much smaller scale MIQP is then solved to find the access points to locally optimize the path.

\subsection{Convex hull formulation}
\noindent First, we demonstrate how to describe the constraint that a UAV stays on a trail using linear equalities and inequalities. Mathematically, this constraint means $x\in \mathbb{R}^2$ belongs to the union of some line segments, where $x$ is the UAV's coordinate.

\begin{figure}[htp]
    \centering
    \includegraphics[width=0.7\linewidth]{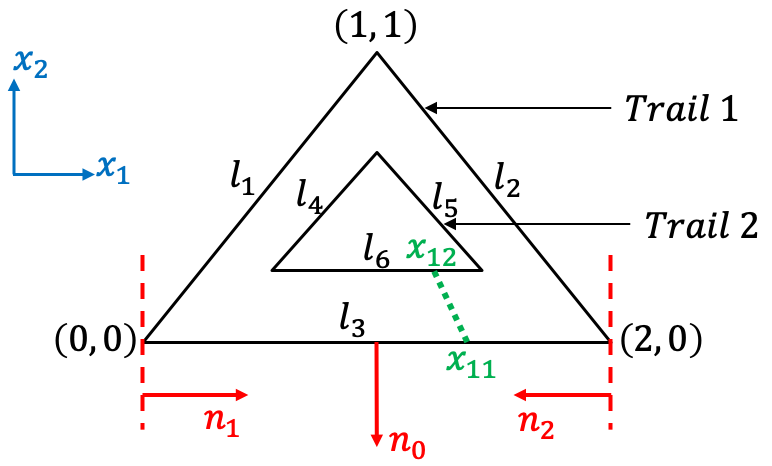}
     \caption{Schematic diagram of trails and their constituent edges. Black triangles represent two trails. $\bm{l}_i$ represent edges, or line segments. $n_0$ is normal to the plane that contains $\bm{l}_3$. $n_1$ and $n_2$ are orthogonal to $n_0$. Numbers in brackets are the coordinates of the vertices of Trail 1.}
     \label{fig:trail_schematics}
\end{figure}

As shown in Fig. \ref{fig:trail_schematics}, a trail consists of the edges of a polygon. Each edge, denoted by $\bm{l}_i$, is a line segment, which can also be written as the following convex set:
\begin{subequations}
\begin{align}
    \bm{l}_i
    &= \{x \in \mathbb{R}^2|\bm{n}_0^T x = a_0, \bm{n}_1^T x \geq a_1, \bm{n}_2^T x \geq a_2\} \label{eqn:trail_definition} \\
    &= \{x \in \mathbb{R}^2 | A_i x \leq b_i \},
\end{align}
\end{subequations}
where Matrix $A_i$ and Vector $b_i$ are the collection of the linear constraints in Eq. (\ref{eqn:trail_definition}).
An example of the constraints that define $\bm{l}_3$ is shown in Fig. \ref{fig:trail_schematics}.

$x$ is on the trail indexed by $j$ and can be formally written as 
\begin{equation}
\label{eqn:point_in_union_of_line_segments}
x \in \bigcup \limits_{i\in I_j} \bm{l}_i,
\end{equation}
where $I_j$ is the set of indices of all edges which belong to Trail $j$. As an example, for the trails shown in Fig. \ref{fig:trail_schematics}, $I_1 = \{1, 2, 3\}$ and $I_2 = \{4, 5, 6\}$.

The constraint stated in Eqn. \ref{eqn:point_in_union_of_line_segments} can be converted to the following mixed-integer implications \cite{valenzuela2016mixed}:
\begin{align}
H_{i} &\Longrightarrow x \in \bm{l}_i, \label{eqn:h_implication} \\
\sum\limits_{i \in I_t} H_i &= 1, \\
H_i &\in \{0,1\}, \forall i \in I_j,
\end{align}
where $H$ is a vector of binary variables. The constraint stated in Eqn. \ref{eqn:h_implication} means that $x$ belongs to Line Segment $\bm{l}_i$ if $H_i = 1$ ($H_i$ is short for $H[i]$, the $i$-th element of $H$). This implication can be further converted to a set of linear constraints using the convex hull formulation \cite{valenzuela2016mixed}:

\begin{align}
x &= \sum_{i \in I_j} x_i, \label{eqn: convex_hull_1}\\
A_i x_i &\leq H_i b_i, \label{eqn: convex_hull_2}\\
H_i x_{lb} &\leq x_i \leq H_i x_{ub},  \label{eqn: convex_hull_3}
\end{align}
where $x_{lb}, x_{ub} \in \mathbb{R}^2$ are the lower and upper bounds of Trail $j$. For instance, $x_{lb} = [0,0]$ and $x_{ub} = [2, 1]$ for Trail 1 in Fig. \ref{fig:trail_schematics}.

\subsection{Full MIQP formulation \label{sec:full_miqp}}
\noindent This sub-section formulates the optimal assignment searching problem defined at the beginning of Sec. \ref{sec:sub_area_planning}, as an MIQP.

The objective of this MIQP is to minimize the total distance traveled by all UAVs in the fleet. Since the optimization has a constraint that all trails must be assigned, the cost function, given by Eqn. \ref{equ:full_miqp_cost}, only needs to account for distances traveled between trails:
\begin{equation}
\underset{\mathbf{x}, H}{\mathrm{min.}}\sum_{k=1}^K\sum_{t=1}^{T-1}\|{\bf x}_{kt}-{\bf x}_{k(t+1)}\|^2.
\label{equ:full_miqp_cost}
\end{equation}

\noindent In Eqn. \ref{equ:full_miqp_cost}, $K \in \mathbb{N}$ is the number of UAVs in the fleet; $T \in \mathbb{N}$ is the planning horizon; $\bf{x}$ is a 3-dimensional array of shape $(K, T, 2)$; and $\mathbf{x}_{kt} \in \mathbb{R}^2$ is a shorthand notation for the slice $\mathbf{x}[k, t]$, which is the coordinate of UAV $k$ at planning Step $t$. $\mathbf{x}_{kt}$ is also the coordinate at which UAV $k$ enters and exits its assigned trail at Step $t$.

The MIQP also needs to satisfy the following constraints:
\begin{align}
&\forall k, t, l, H_{ktl} \Longrightarrow x_{kt} \in \bm{l}_l, H_{ktl}\in\{0,1\}, \label{equ:implication}\\
&\forall k, t, \sum_{l=1}^LH_{ktl} = 1, \label{equ:unique_existence}\\
&\forall i, \sum_{k=1}^K\sum_{t=1}^T\sum_{l \in I_i}H_{ktl} = 1, \label{equ:unique_visit} \\
&\forall k, \sum_{t=1}^T\sum_{i=1}^{N_t}C_i\sum_{l \in I_i} H_{ktl} \leq D,
\label{equ:battery}
\end{align}
where $N_t$ is the total number of trails; $L$ is the total number of line segments ($\bm{l}$'s) in all trails; $I_i$ is the set of line segments indices of Trail $i$; $C_i$ is the perimeter of Trail $i$; $D$ is the maximum distance a UAV can travel with one battery charge; $H$ is a 3-dimensional binary array of shape $(K, T, L)$.

Constraint (\ref{equ:implication}) means if $H_{ktl}=1$, UAV $k$ is on Line Segment $l$ at Step $t$. It can be expanded into linear constraints using Eqn. \ref{eqn: convex_hull_1} to \ref{eqn: convex_hull_3}. Constraint (\ref{equ:unique_existence}) means each UAV cannot appear on more than one line segment at each planning step. Constraint (\ref{equ:unique_visit}) guarantees that each trail is assigned exactly once to one UAV. Eqn. \ref{equ:battery} ensures that each UAV can traverse all of its assigned trails without changing batteries. 
After obtaining the optimal solution, the plan can be constructed from the solution as follows:
\begin{itemize}
\item Trail $i$ is assigned to UAV {k} if $\sum_{t=1}^T \sum_{l \in I_i} H_{ktl} = 1$. Looping through all $k$ and $I_i$ recovers the sequence of trails assigned to each UAV. 

\item For each UAV, the entrance/exit point of each of its assigned trail can be calculated using Eqn. (\ref{eqn: convex_hull_1}).
\end{itemize}

The MIQP formulated in this sub-section has $K\times T \times L$ binary variables, which quickly becomes intractable even for moderately-sized problems: the planner simply has too many decisions to make. To reduce the number of binary variables, we decompose the planning problem into two stages: 
\begin{enumerate}
\item A relative good trail assignment is obtained using a combination of Random Key Genetic Algorithm (RKGA) and Modified Vehicle Routing Problem (MVRP). (detailed in Sub-sec. \ref{sec:RKGA-MVRP})
\item A smaller MIQP is solved to optimize access points over the fixed trail assignment. (detailed in Sub-sec. \ref{sec:small_MIQP})
\end{enumerate}

\noindent Although global optimality is sacrificed, the proposed two-step optimization approach has proven to generate good enough plans within a reasonable amount of time.

\subsection{Finding good trail assignment with RKGA and MVRP \label{sec:RKGA-MVRP}}

\noindent Genetic Algorithms are used to search for "optimal" solutions by \textit{evolving} a set of feasible solutions, or a \textit{population} of \textit{chromosomes}, until the maximum generation achieved, or the termination condition is meet. 

\subsubsection{Chromosomes}
\begin{figure}[htp!]
	\centering 	
    \includegraphics[width=0.56\linewidth]{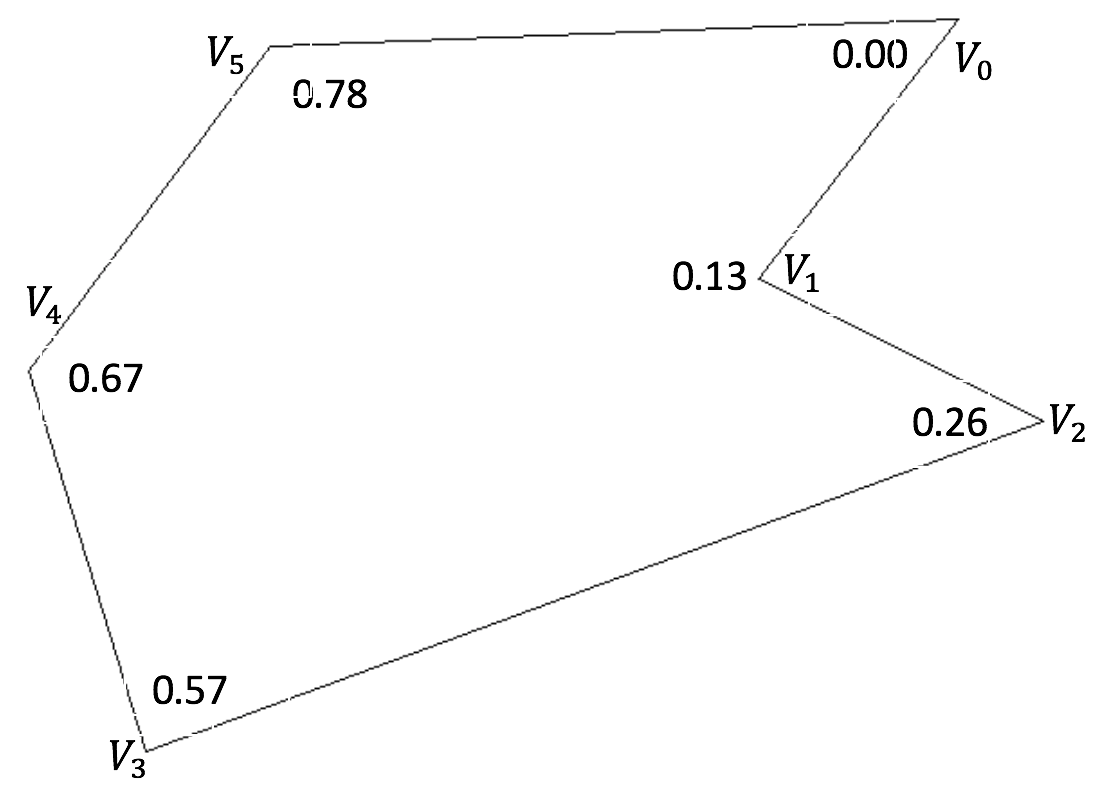}
	\caption{A trail and the encoded values of its vertices.}
	\label{fig:encode}
\end{figure}

To solve the trail assignment problem, we structure the population as a matrix, $P \in [0, 1)^{N \times N_t}$, where $N$ is the number of chromosomes, and $N_t$ the total number of trails in a map. $P_i$, the $i$-th row of $P$, is a chromosome. $P_i[j] \in [0, 1)$ can be mapped from an access point on Trail $j$ using the following encoder function ($E: \mathbb{R}^2 \rightarrow [0, 1)$):
\begin{equation}
E(x) = 
\begin{cases}
\|x - v_{0}\|_2/C, x \in \overline{v_{0}v_{1}} \\
E(v_{k}) + \|x - v_{k}\|_2/C, x \in \overline{v_{k}v_{k+1}}, k \geq 1, \\
\end{cases}
\label{eqn:encoder}
\end{equation}
where $x \in \mathbb{R}^2 $ is the coordinate of a point on a trail, $v_i \in \mathbb{R}^2$ the coordinates of the vertices of the trail, $\overline{v_{k}v_{k+1}}$ the line segment between $v_k$ and $v_{k+1}$, and $C$ the perimeter of the trail. $E(x)$ represents the normalized distance of $x$ from $v_{0}$ measured along the perimeter of the trail. The encoding allows generating feasible access point candidates via random sampling \cite{jing2017sampling}. An example of such encoding is shown in Fig. \ref{fig:encode}.

The decoder function ($D: [0, 1) \rightarrow \mathbb{R}^2 $ ) is the inverse of $E$:
\begin{equation}
D(p) = 
v_{k} + C(p - E(v_k))\frac{v_{k+1} - v_k}{\|v_{k+1}-v_k\|_2},
\label{eqn:decoder}
\end{equation}
where $p$ is the encoded value of Point $x$, and $k$ is such that $E(v_k) \leq p \leq E(v_{k+1})$.

\begin{algorithm}[htp!]
\caption{Evaluate fitness of a population}
\label{algo:fitness}
\begin{algorithmic}[1]
\Function{EvaluateFitness}{$P$, $N$, $N_t$, $K$, Trails}
\State fitness = []
\State assignments = []
\For{j$<$N}
\State X $\leftarrow$ Decode($P_j$, Trails)
\State [assignment, tourLength]$ \leftarrow$ MVRP(X, $K$, Trails)
\State fitness.append(tourLength)
\State assignments.append(assignment)
\EndFor
\State [$P$, fitness, assignments] $\leftarrow$ Sort($P$, fitness, assignments)
\State \Return $P$, fitness, assignments
\EndFunction
\end{algorithmic}
\end{algorithm}

\subsubsection{Evaluating fitness of a population}
The function defined in Alg. \ref{algo:fitness} evaluates the fitness of every chromosome $P_j$ in a population $P$. Trails in Line 1 refers to the coordinates and encoded values of the vertices of all trails.

Each $P_j$ is first decoded into 2D coordinates of access points of all trails by repeatedly calling Eqn. \ref{eqn:decoder} (Line 5). After the access points are fixed, the problem of assigning trails to a fleet of $K$ UAVs is equivalent to a variant of the Vehicle Routing Problem (VRP) with capacity constraints and arbitrary start and end points \cite{yuan2017towards}, which we term as the Modified VRP (MVRP). The input to the MVRP is a fully-connected graph whose nodes are made up by $X$, the decoded chromosome. Accordingly, the fitness of a chromosome can be defined as the length of the longest tour (tourLength in Line 6), which can be interpreted as the maximum flight distance among all UAVs. The MVRP can be efficiently solved by an open source combinatorial optimization software called OR-Tools developed by Google AI \cite{ortool}. The function call to solve MVRP (Line 6) returns the optimal assignment of trails to UAVs together with the fitness of $P_j$. Lastly, the population and the corresponding trail assignments are sorted in ascending order by their fitness values (Line 10).

\begin{algorithm}[htp!]
\caption{MVRP-RKGA}\label{algo:MVRP-RKGA}
\begin{algorithmic}[1]
\Require $N$, $N_t$, $K$, Trails, parentSelectNum\par 
\hskip 0pt crossoverRate, mutateRate, eliteNum, \par
\hskip 0pt offspringNum, iterationNum
\State $P$ $\leftarrow$ UniformSample($N$, $N_t$)
\State [$P$, fitness] $\leftarrow$  EvaluateFitness($P$, $N$, $N_t$, $K$, Trails)
\For{iteration $<$ iterationNum}
\State parents$\leftarrow$\\\hspace{0.5cm}SelectParent($P$,parentSelectNum,offspringNum)
\State offspring $\leftarrow$ CrossOver(parents, crossoverRate)
\State offspring $\leftarrow$ Mutate(offspring, mutateRate)
\State $P_{new}$ = Merge($P_{0:eliteNum}$, offspring)
\State [$P$, fitness, assignments] $\leftarrow$ EvaluateFitness($P_{new}$, $N$, $N_t$, $K$, Trails)
\EndFor
\State \Return $P_0$, assignments[0]
\end{algorithmic}
\end{algorithm}

\subsubsection{Evolving the population}
Alg. \ref{algo:MVRP-RKGA} evolves a population using Genetic Algorithm, with fitness of chromosomes evaluated by Alg. \ref{algo:fitness}. Firstly, a population is initialized by uniform sampling between 0 and 1 (Line 1). The population's fitness is also initialized (Line 2). Secondly, the population is evolved using the classical genetic algorithm: parents are randomly selected (Line 4); off-springs are created by cross-over (Line 5) and mutation (Line 6); and the next-generation population ($P_{new}$) is created by combining current-gen elites and off-springs (Line 7). After a fixed number of iterations, the fittest chromosome $P_0$ and its UAV assignments are returned (Line 10).

\subsection{Reduced MIQP for access points optimization \label{sec:small_MIQP}}
\noindent With a fixed trail assignment returned by Alg. \ref{algo:MVRP-RKGA}, a smaller MIQP with $L$ binary variables can be formulated to locally optimize the access points. The objective of this MIQP is given by:
\begin{equation}
\sum_{k=1}^K \sum_t \|\mathbf{x}_{O(k, t+1)} - \mathbf{x}_{O(k, t)}\|^2,
\end{equation}
 where $\mathbf{x} \in \mathbb{R}^{2 \times N_t}$, $\mathbf{x}_i$ (Row $i$ of $\mathbf{x}$) is the access point of Trail $i$, $O(k, t)$ is a function that returns the index of the trail assigned to UAV $k$ at Step $t$. This function is completely defined by a given trail assignment.

The following constraints confine $\mathbf{x}_i$ to Trail $i$:  
\begin{align}
&\forall l \in I_i, H_l = 1 \Longrightarrow \mathbf{x}_i \in \bm{l}_l, \\
&\forall 1 \leq i \leq N_t, \sum_{l \in I_i} H_l = 1.
\end{align}
\begin{figure}[htp!]
	\centering 	
\includegraphics[width=0.98\linewidth]{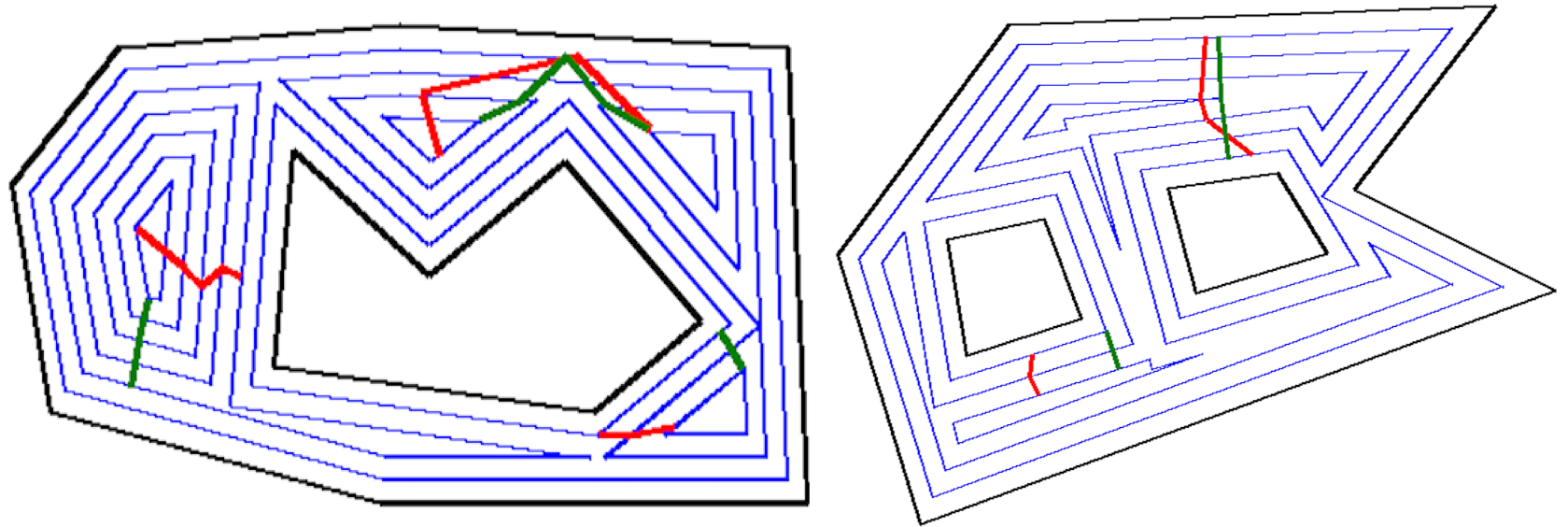}
	\caption{Path between polygonal trails with MVRP-RKGA and MIQP. Red lines are the paths calculated from MVRP-RKGA and green lines are the paths optimized with MIQP.}
	\label{fig:miqp_result}
\end{figure}

\noindent Although global optimality is not guaranteed, the local optimization can still make a significant improvement over the feasible solution returned by Alg. \ref{algo:MVRP-RKGA}. The MIQP problem is solved with Drake \cite{drake}, an open-source optimization toolbox with interface to Python. As shown in Fig. \ref{fig:miqp_result}, MIQP optimizes the positions of access points obtained by MVRP-RKGA to reduce inter-trail distances by 37\%(left) and 12\%(right). Moreover, it only takes the solver 0.09s (left) and 0.83s (right) to find these locally optimal solutions.

\section{Car Routing}
\begin{figure}[!htp]
	\centering 
    \includegraphics[width=0.55\linewidth]{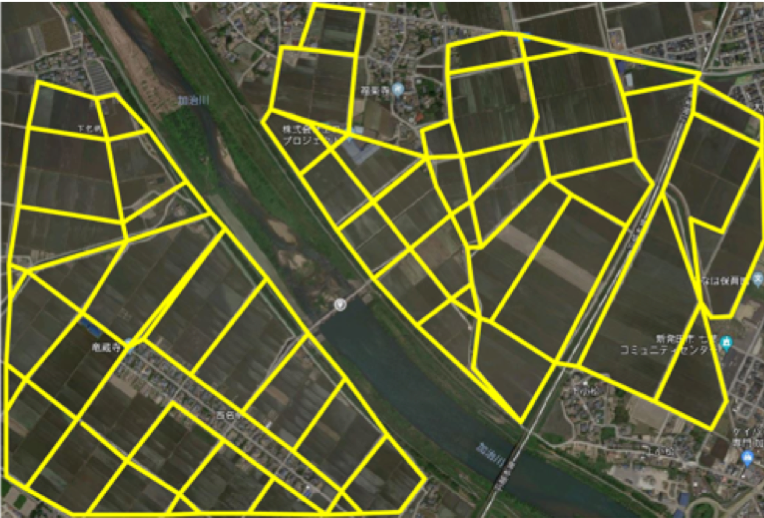}
	\caption{Road extracted from a map}
	\label{fig:road_extraction}
\end{figure}

\noindent GPS information of roads around and inside the farmlands is extracted from OpenStreetMap \cite{OpenStreetMap}, as shown in Fig. \ref{fig:road_extraction}.

The extracted road network is represented as a graph $G_{road}=(V_{road},E_{road},W_{road})$, where $v\in V_{road}$ represents parking spots, $e\in E_{road}$ denotes road segments and $W_{road}$ is the distance of the road segment connecting two intersections.

\subsection{Car routing within sub-area}
\noindent After identifying the routes for UAVs, we will assign UAVs' take-off and landing spot (car location) in each sub-area to minimize the total flight distance between start/end access points and the distance traveled by the ground vehicle (car):
\begin{equation}
\underset{P_{car}}{\mathrm{min.}}\sum_{k=1}^K||P_{car}-P_{uav(k)}||^2,
\end{equation}
where $P_{car}$ is the position of the ground vehicle and $P_{uav(k)}$ is the position of the $k_{th}$ UAV.

Assuming that the car can only dispatch and receive UAVs at $V_{road}$ in each sub-area and yellow dots represent all possible positions of the car. Therefore, the route of the car within the sub-area is the shortest path from a red dot to a green dot along the weighted road graph. 

\begin{figure}[htp]
	\centering 	
    \includegraphics[width=0.33\linewidth]{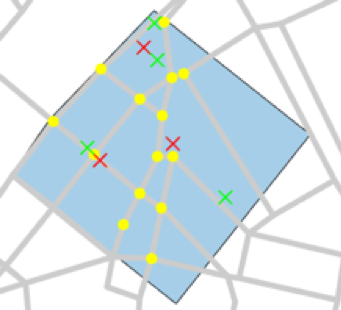} 
    \includegraphics[width=0.3\linewidth]{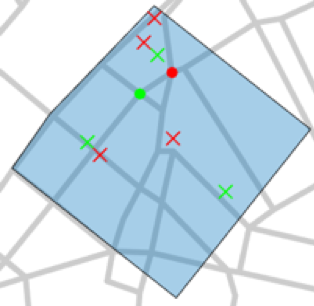}
    \includegraphics[width=0.3\linewidth]{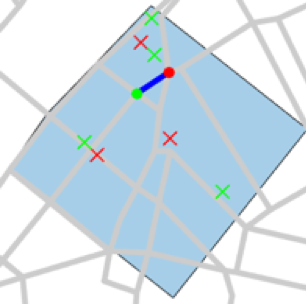}
	\caption{UAVs' release and landing spots on road. '$\textcolor{red}{\times}$' and '$\textcolor{green}{\times}$' denote the start and end access points of UAVs in the planned trails, while red and green dots are the chosen take-off and landing positions for UAVs. The blue line is the car route from start to end position.}
	\label{fig:uav_setoff_subgraph}
\end{figure}
\subsection{Car routing between sub-areas}
\noindent Given the start and end locations of the car in each sub-area, we would like to find the shortest path of the car to visit all sub-areas. As the start and end positions of the car in each sub-area sometimes are different, this problem is formulated as a Asymmetric-cost Traveling Salesman's Problems (ATSP), which can be solved with Google OR-Tools \cite{ortool}. Each sub-area is considered as a node, while the distance from Sub-area $i$ to Sub-area $j$ equals to the length of shortest path from the final car position in Sub-area $i$ to the start car position in Sub-area $j$.

\section{RESULTS AND DISCUSSION \label{sec:result}}
\noindent We use the following set of UAV specifications based on the agriculture drone T16 released by DJI \cite{wang2019applications} when generating the numerical results in this section. Each UAV has a flight endurance of 10 minutes. All UAVs cruise at a speed of 6m/s with 6.5m coverage width. During each take-off and landing cycle, 10 minutes is spent on spraying pesticides and 5 minutes on traveling between the farmlands and a ground vehicle (car). The car only releases and picks up UAVs at intersections of roads in the given map. Furthermore, the car must always stay within the transmission distance of all UAVs in this example we set 500m. 

\begin{figure}[htp!]
    \vspace{5pt}
    \centering
    \includegraphics[width=0.85\linewidth]{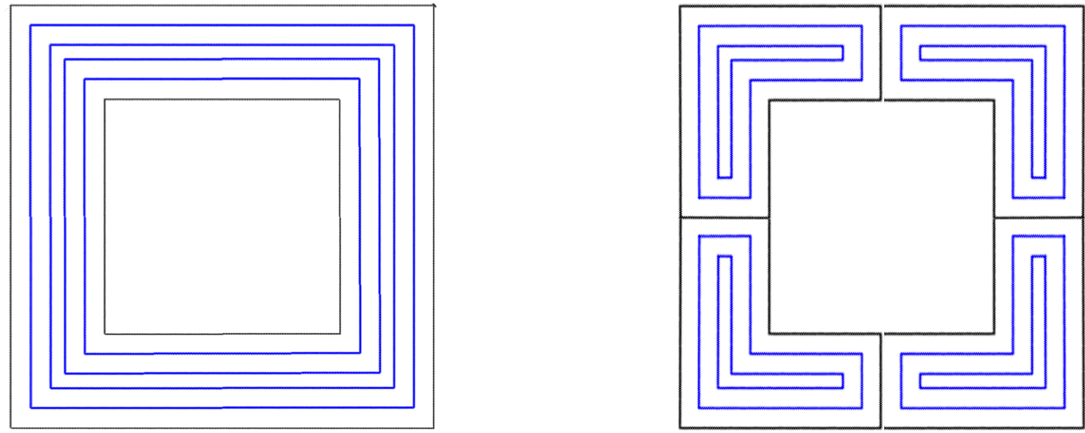}
    \caption{Path before vs. after partition for each UAV. Blue lines are the path for UAVs.}
    \label{fig:path_partition}
\end{figure}

\begin{figure}[htp]
	\centering 	
\includegraphics[width=0.46\linewidth]{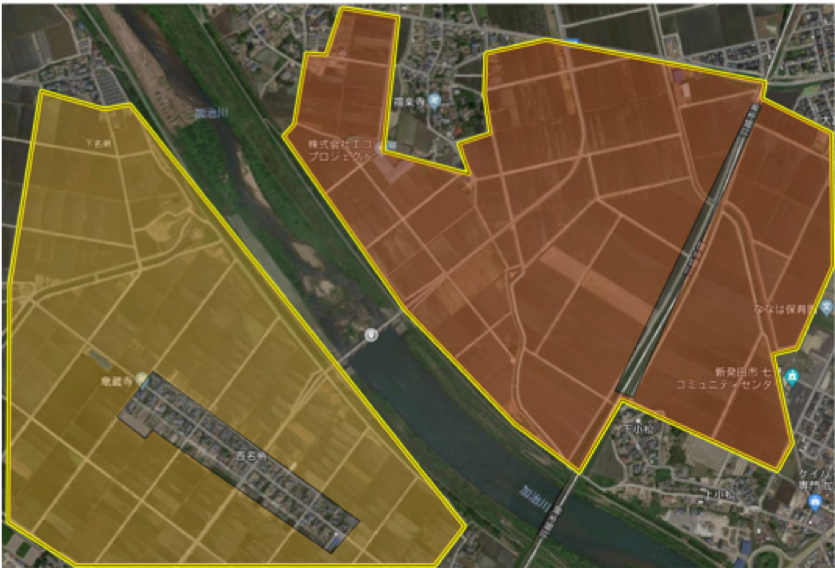}
\includegraphics[width=0.46\linewidth]{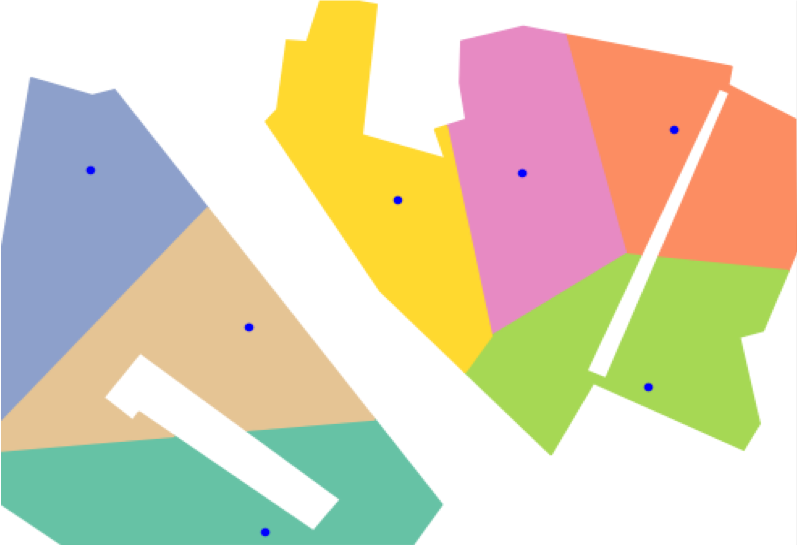}
	\caption{Farm partitioning result. Different colors represent different sub-areas and white color areas are obstacles.}
	\label{fig:farm_partition}
\end{figure}

We tested the area partitioning algorithm with multiple maps based on the maximum coverage area for a fleet of UAVs and the maximum communication distance between the UAVs and the car. Compared with the most common approach of assigning sub-areas to UAVs \cite{karapetyan2017efficient, karapetyan2018multi, araujo2013multiple, barrientos2011aerial, balampanis2017spiral}, our approach reduces the total number of turns from 48 turns to 16 turns as demonstrated in Fig. \ref{fig:path_partition} for the mitered-offset path. The partition in Fig. \ref{fig:farm_partition} was generated using a population of 200 chromosomes, and 15 iterations takes 7s on a 2.7 GHz Intel Core i5 laptop.

\begin{figure}[htp]
	\centering 	
    \includegraphics[width=0.85\linewidth]{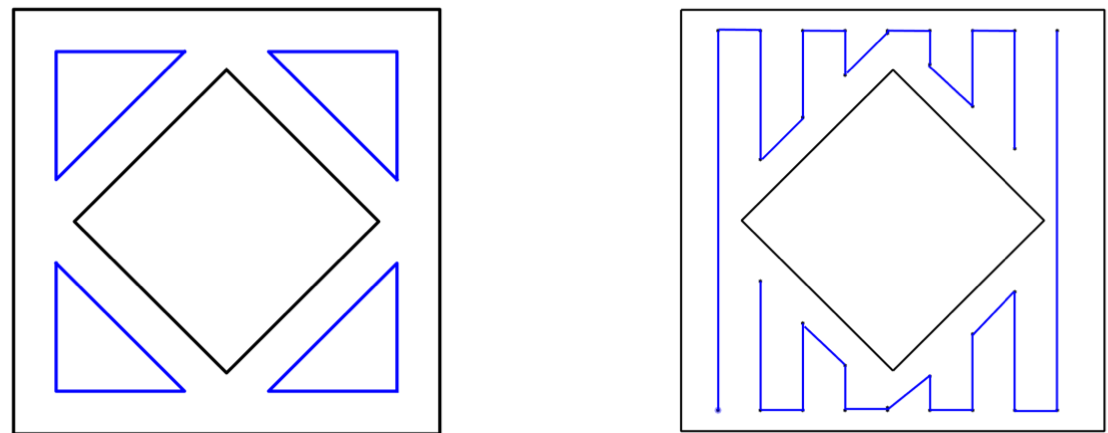}
	\caption{Mitered offset path vs. Zigzag path. Blue lines are the path for UAVs.}
	\label{fig:path_pattern_comparison}
\end{figure}

The next step is to generate paths within sub-areas. In many situations, paths generated by mitered-offset are shorter, have less turning angles and total number of turns than zig-zag paths. (Table \ref{tab:zigVSloop} and Fig. \ref{fig:path_pattern_comparison}).

\begin{table}[h!]
\begin{center}
\caption{Comparison between mitered offset path and zigzag path}
\label{tab:zigVSloop}
\begin{tabular}{c c c}
                    & \textbf{offset} & \textbf{zigzag}   \\
                     \hline
path length (m)        & 2881 &  3517  \\
number of turns & 12 & 24 \\
total turning angle & 4$\pi$   & 12$\pi$  \\
\hline
\end{tabular}
\end{center}
\vspace{-10pt}
\end{table}

\begin{figure}[!htp]
    \vspace{5pt}
	\centering 	
    \includegraphics[width=0.98\linewidth]{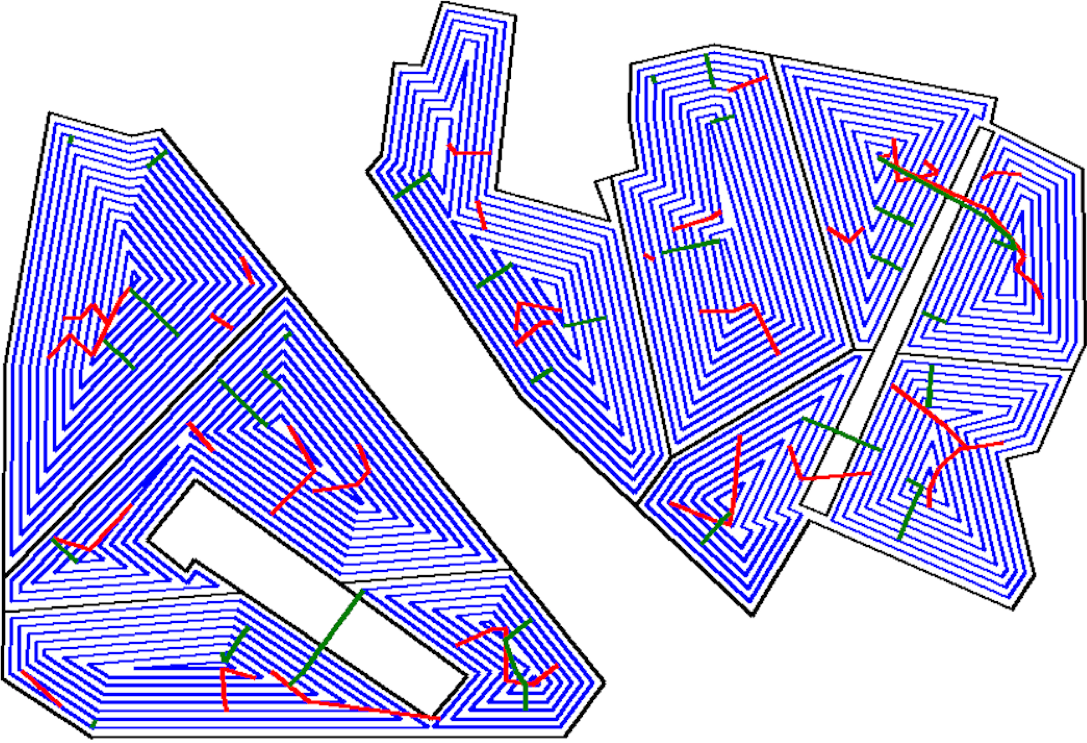}
	\caption{Planned paths for UAVs in fields with MVRP-RKGK and MIQP. Red lines represent the UAV paths between trails generated by MVRP-RKGA. Green lines denote the paths improved with MIQP.}
	\label{fig:ga_result_all}
\end{figure}

\begin{figure}[h!]
	\centering 	
    \includegraphics[width=0.95\linewidth]{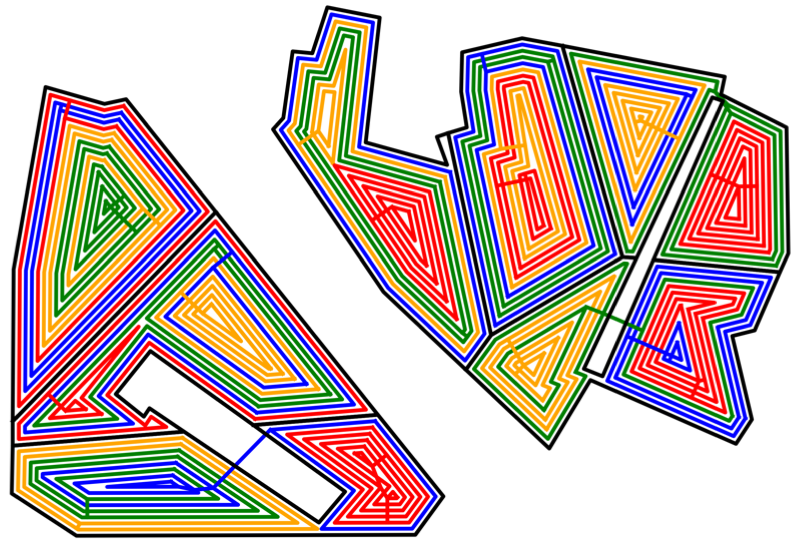}
	\caption{Trail assignments generated by MVRP-RKGA100. Trails with the same color are assigned to the same UAV. Trails assigned to the same UAV are connected by paths optimized by MIQP.}
	\label{fig:miqp_result_all}
\end{figure}

Next, in each sub-area, mitered-offset trails are assigned to UAVs using MVRP-RKGA and MIQP. The trail assignments generated by MVRP-RKGA100 (MVRP-RKGA with a population of 100) are shown in Fig. \ref{fig:miqp_result_all}. The access points generated by MVRP-RKGA100 are shown as connected by red line segments in Fig.  \ref{fig:ga_result_all}. The access points improved by the MIQP in Sub-sec. \ref{sec:small_MIQP} are shown in the same figure, connected by green lines. Compared with MVRP-RKGA100, access points further optimized by MIQP can reduce flying distances between trails by up to 50\%, as shown in Table \ref{tab:1500VS1001}. The total computation time of MVRP-RKGA100 and MIQP is less than 4 minutes when running on a 2.7 GHz Intel Core i5 laptop. 

\begin{table}[h!]
\begin{center}
\caption{Planning time and flight distance of MVRP-RKGA100 and MIQP}
\label{tab:1500VS1001}
\begin{tabular}{c |c c c c c c c}
                 \textbf{time}   & \textbf{1} & \textbf{2} & \textbf{3} & \textbf{4}& \textbf{5} & \textbf{6}& \textbf{7}   \\
                     \hline
GA100	(s)&34 & 51& 26&14 & 43 & 15 &17\\
MIQP (s)	&0.28 & 0.18 & 2 & 9.5 & 13 & 2.6 & 1.0\\
\hline
\textbf{distance} \\
\hline
GA100(m)& 841 & 761 & 415 & 424& 777 & 341 & 648\\
MIQP(m) &532 & 259 & 244 & 202 & 432& 186&  228\\
\hline
\end{tabular}
\end{center}
\end{table}

\begin{figure}[!htp] 
    \vspace{8pt}
	\centering 	
    \includegraphics[width=0.8\linewidth]{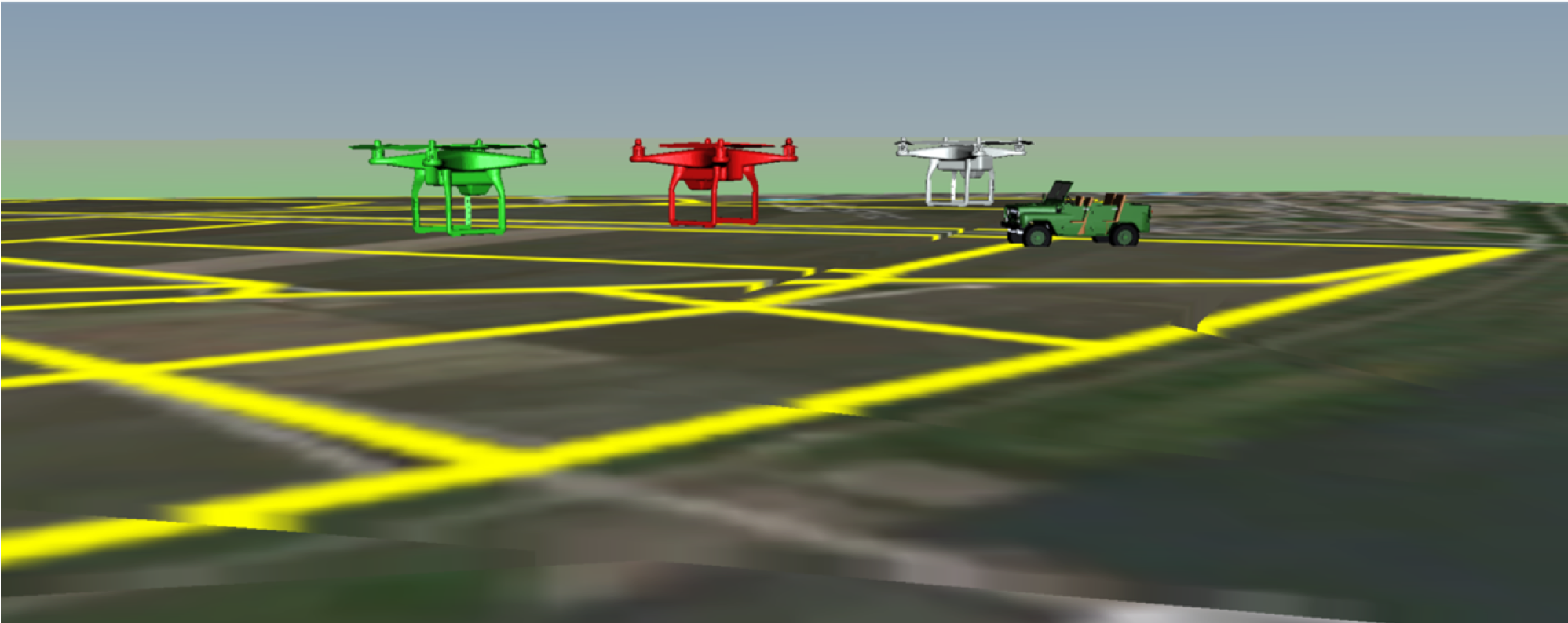}
	\caption{Simulation environment}
	\label{fig:3d_simulation}
\end{figure}

\begin{figure}[!htp] 
	\centering 	
    \includegraphics[width=0.6\linewidth]{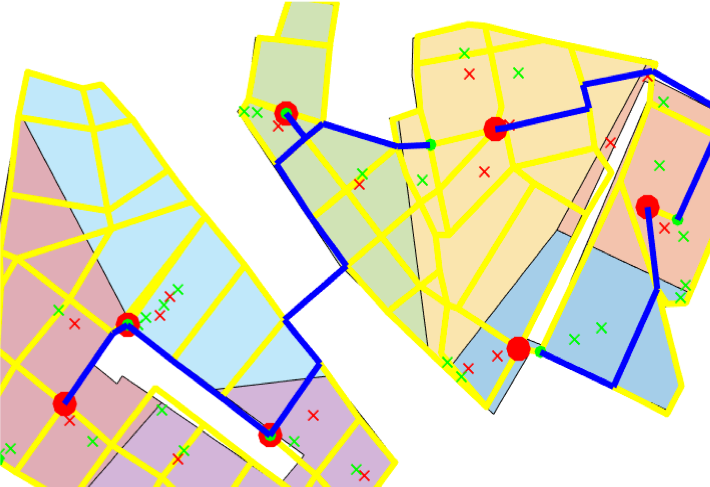} 
	\caption{Car routing between sub-areas}
	\label{fig:car_routing}
\end{figure}

Execution of a full plan for a fleet of four UAVs and one car, including the planned UAV paths in Fig. \ref{fig:miqp_result_all} and the planned car paths in Fig. \ref{fig:car_routing}, is simulated in the Simulink-based 3D environment shown in Fig. \ref{fig:3d_simulation}. The heterogeneous fleet successfully covers the given farmland. The operation time in each sub-area is illustrated in Table \ref{tab:1500VS1003}. Assuming the time for swapping battery and the car traveling between sub-areas takes 10 min, it takes 1.5 hours to cover the farm with an area of 617,210 $m^2$. 

\begin{table}[h!]
\begin{center}
\caption{Time consumption for heterogeneous vehicles to cover sub-areas }
\label{tab:1500VS1003}
\begin{tabular}{c |c c c c c c c}
                    & \textbf{1} & \textbf{2} & \textbf{3} & \textbf{4}& \textbf{5} & \textbf{6}& \textbf{7}   \\
                     \hline
Flight time (min)&8.97&    9.97   & 9.39  &  8.53   & 8.44 &  9.91  &  9.34\\
\hline
\end{tabular}
\end{center}
\vspace{-15pt}
\end{table}
\section*{ACKNOWLEDGMENT}
\noindent The authors would like to thank TOPRISE Co., LTD. for their sponsorship for this work, 
Mayuko Nemoto for help on problem formulation and Runlin Duan and Kuangjie Sheng for running experiments.
\bibliographystyle{IEEEtran}
\bibliography{main}

\begin{thebibliography}{10}
\providecommand{\url}[1]{#1}
\csname url@samestyle\endcsname
\providecommand{\newblock}{\relax}
\providecommand{\bibinfo}[2]{#2}
\providecommand{\BIBentrySTDinterwordspacing}{\spaceskip=0pt\relax}
\providecommand{\BIBentryALTinterwordstretchfactor}{4}
\providecommand{\BIBentryALTinterwordspacing}{\spaceskip=\fontdimen2\font plus
\BIBentryALTinterwordstretchfactor\fontdimen3\font minus
  \fontdimen4\font\relax}
\providecommand{\BIBforeignlanguage}[2]{{%
\expandafter\ifx\csname l@#1\endcsname\relax
\typeout{** WARNING: IEEEtran.bst: No hyphenation pattern has been}%
\typeout{** loaded for the language `#1'. Using the pattern for}%
\typeout{** the default language instead.}%
\else
\language=\csname l@#1\endcsname
\fi
#2}}
\providecommand{\BIBdecl}{\relax}
\BIBdecl

\bibitem{xiongkui2017recent}
H.~Xiongkui, J.~Bonds, A.~Herbst, and J.~Langenakens, ``Recent development of
  unmanned aerial vehicle for plant protection in east asia,''
  \emph{International Journal of Agricultural and Biological Engineering},
  vol.~10, no.~3, pp. 18--30, 2017.

\bibitem{dji}
\BIBentryALTinterwordspacing
DJI, ``Agrasmg-1,'' 2018. [Online]. Available: \url{https://www.dji.com/mg-1}
\BIBentrySTDinterwordspacing

\bibitem{karapetyan2018multi}
N.~Karapetyan, J.~Moulton, J.~S. Lewis, A.~Q. Li, J.~M. O'Kane, and
  I.~Rekleitis, ``Multi-robot dubins coverage with autonomous surface
  vehicles,'' in \emph{2018 IEEE International Conference on Robotics and
  Automation (ICRA)}.\hskip 1em plus 0.5em minus 0.4em\relax IEEE, 2018, pp.
  2373--2379.

\bibitem{jing2016view}
W.~Jing, J.~Polden, P.~Y. Tao, W.~Lin, and K.~Shimada, ``View planning for 3d
  shape reconstruction of buildings with unmanned aerial vehicles,'' in
  \emph{International Conference on Control, Automation, Robotics and
  Vision}.\hskip 1em plus 0.5em minus 0.4em\relax IEEE, 2016, pp. 1--6.

\bibitem{faiccal2017adaptive}
B.~S. Fai{\c{c}}al, H.~Freitas, P.~H. Gomes, L.~Y. Mano, G.~Pessin, A.~C.
  de~Carvalho, B.~Krishnamachari, and J.~Ueyama, ``An adaptive approach for
  uav-based pesticide spraying in dynamic environments,'' \emph{Computers and
  Electronics in Agriculture}, vol. 138, pp. 210--223, 2017.

\bibitem{galceran2013survey}
E.~Galceran and M.~Carreras, ``A survey on coverage path planning for
  robotics,'' \emph{Robotics and Autonomous systems}, vol.~61, no.~12, pp.
  1258--1276, 2013.

\bibitem{choset2001coverage}
H.~Choset, ``Coverage for robotics--a survey of recent results,'' \emph{Annals
  of mathematics and artificial intelligence}, vol.~31, no. 1-4, pp. 113--126,
  2001.

\bibitem{choset2000coverage}
------, ``Coverage of known spaces: The boustrophedon cellular decomposition,''
  \emph{Autonomous Robots}, vol.~9, no.~3, pp. 247--253, 2000.

\bibitem{maza2007multiple}
I.~Maza and A.~Ollero, ``Multiple uav cooperative searching operation using
  polygon area decomposition and efficient coverage algorithms,'' in
  \emph{Distributed Autonomous Robotic Systems 6}.\hskip 1em plus 0.5em minus
  0.4em\relax Springer, 2007, pp. 221--230.

\bibitem{balampanis2017area}
F.~Balampanis, I.~Maza, and A.~Ollero, ``Area partition for coastal regions
  with multiple uas,'' \emph{Journal of Intelligent \& Robotic Systems},
  vol.~88, no. 2-4, pp. 751--766, 2017.

\bibitem{bast2000area}
H.~Bast and S.~Hert, ``The area partitioning problem,'' 2000.

\bibitem{bormann2018indoor}
R.~Bormann, F.~Jordan, J.~Hampp, and M.~H{\"a}gele, ``Indoor coverage path
  planning: Survey, implementation, analysis,'' in \emph{2018 IEEE
  International Conference on Robotics and Automation (ICRA)}.\hskip 1em plus
  0.5em minus 0.4em\relax IEEE, 2018, pp. 1718--1725.

\bibitem{isler2018coverage}
V.~Isler and M.~Wei, ``Coverage path planning under the energy constraint,''
  2018.

\bibitem{richards2015user}
D.~Richards, T.~Patten, R.~Fitch, D.~Ball, and S.~Sukkarieh, ``User interface
  and coverage planner for agricultural robotics,'' in \emph{Proceedings of
  ARAA Australasian conference on robotics and automation (ACRA). Google
  Scholar}, 2015.

\bibitem{moravec1985high}
H.~Moravec and A.~Elfes, ``High resolution maps from wide angle sonar,'' in
  \emph{Proceedings. 1985 IEEE international conference on robotics and
  automation}, vol.~2.\hskip 1em plus 0.5em minus 0.4em\relax IEEE, 1985, pp.
  116--121.

\bibitem{barrientos2011aerial}
A.~Barrientos, J.~Colorado, J.~d. Cerro, A.~Martinez, C.~Rossi, D.~Sanz, and
  J.~Valente, ``Aerial remote sensing in agriculture: A practical approach to
  area coverage and path planning for fleets of mini aerial robots,''
  \emph{Journal of Field Robotics}, vol.~28, no.~5, pp. 667--689, 2011.

\bibitem{brown2017coverage}
S.~Brown, ``Coverage path planning and room segmentation in indoor environments
  using the constriction decomposition method,'' Master's thesis, University of
  Waterloo, 2017.

\bibitem{jing2016sampling}
W.~Jing, J.~Polden, W.~Lin, and K.~Shimada, ``Sampling-based view planning for
  3d visual coverage task with unmanned aerial vehicle,'' in \emph{IEEE/RSJ
  International Conference on Intelligent Robots and Systems (IROS)}.\hskip 1em
  plus 0.5em minus 0.4em\relax IEEE, 2016, pp. 1808--1815.

\bibitem{maini2018cooperative}
P.~Maini, K.~Sundar, S.~Rathinam, and P.~Sujit, ``Cooperative planning for
  fuel-constrained aerial vehicles and ground-based refueling vehicles for
  large-scale coverage,'' \emph{arXiv preprint arXiv:1805.04417}, 2018.

\bibitem{deng2018heterogeneous}
D.~Deng, P.~Palli, F.~Shu, K.~Shimada, and T.~Pang, ``Heterogeneous vehicles
  routing for water canal damage assessment,'' in \emph{2018 IEEE/RSJ
  International Conference on Intelligent Robots and Systems (IROS)}.\hskip 1em
  plus 0.5em minus 0.4em\relax IEEE, 2018, pp. 2375--2382.

\bibitem{huber2011computing}
S.~Huber, \emph{Computing straight skeletons and motorcycle graphs: theory and
  practice}.\hskip 1em plus 0.5em minus 0.4em\relax Shaker, 2011.

\bibitem{valenzuela2016mixed}
A.~K. Valenzuela, ``Mixed-integer convex optimization for planning aggressive
  motions of legged robots over rough terrain,'' Ph.D. dissertation,
  Massachusetts Institute of Technology, 2016.

\bibitem{jing2017sampling}
W.~Jing, J.~Polden, C.~F. Goh, M.~Rajaraman, W.~Lin, and K.~Shimada,
  ``Sampling-based coverage motion planning for industrial inspection
  application with redundant robotic system,'' in \emph{Intelligent Robots and
  Systems (IROS), 2017 IEEE/RSJ International Conference on}.\hskip 1em plus
  0.5em minus 0.4em\relax IEEE, 2017, pp. 5211--5218.

\bibitem{yuan2017towards}
B.~Yuan and T.~Zhang, ``Towards solving tspn with arbitrary neighborhoods: A
  hybrid solution,'' in \emph{Australasian Conference on Artificial Life and
  Computational Intelligence}.\hskip 1em plus 0.5em minus 0.4em\relax Springer,
  2017, pp. 204--215.

\bibitem{ortool}
\BIBentryALTinterwordspacing
G.~AI, ``Google optimization tools,'' 2018. [Online]. Available:
  \url{https://developers.google.com/optimization/}
\BIBentrySTDinterwordspacing

\bibitem{drake}
\BIBentryALTinterwordspacing
R.~Tedrake and the Drake Development~Team, ``Drake: A planning, control, and
  analysis toolbox for nonlinear dynamical systems,'' 2016. [Online].
  Available: \url{https://drake.mit.edu}
\BIBentrySTDinterwordspacing

\bibitem{OpenStreetMap}
{OpenStreetMap contributors}, ``{Planet dump retrieved from
  https://planet.osm.org },'' \url{ https://www.openstreetmap.org }, 2017.

\bibitem{wang2019applications}
L.~Wang, Y.~Lan, Y.~Zhang, H.~Zhang, M.~N. Tahir, S.~Ou, X.~Liu, and P.~Chen,
  ``Applications and prospects of agricultural unmanned aerial vehicle obstacle
  avoidance technology in china,'' \emph{Sensors}, vol.~19, no.~3, p. 642,
  2019.

\bibitem{karapetyan2017efficient}
N.~Karapetyan, K.~Benson, C.~McKinney, P.~Taslakian, and I.~Rekleitis,
  ``Efficient multi-robot coverage of a known environment,'' in
  \emph{Intelligent Robots and Systems (IROS), 2017 IEEE/RSJ International
  Conference on}.\hskip 1em plus 0.5em minus 0.4em\relax IEEE, 2017, pp.
  1846--1852.

\bibitem{araujo2013multiple}
J.~Araujo, P.~Sujit, and J.~B. Sousa, ``Multiple uav area decomposition and
  coverage,'' in \emph{2013 IEEE symposium on computational intelligence for
  security and defense applications (CISDA)}.\hskip 1em plus 0.5em minus
  0.4em\relax IEEE, 2013, pp. 30--37.

\bibitem{balampanis2017spiral}
M.~I. Balampanis, Fotios and A.~Ollero, ``Spiral-like coverage path planning
  for multiple heterogeneous uas operating in coastal regions,'' in
  \emph{Unmanned Aircraft Systems (ICUAS), 2017 International Conference
  on}.\hskip 1em plus 0.5em minus 0.4em\relax IEEE, 2017, pp. 617--624.

\end{thebibliography}
\end{document}